\documentclass[runningheads]{llncs}
\usepackage{graphicx}
\usepackage{comment}
\usepackage{amsmath,amssymb} 
\usepackage{color}
\usepackage{multirow}

\begin{document}
\pagestyle{headings}
\mainmatter
\def\ECCVSubNumber{3823}  

\title{Dense Semantic 3D Map Based Long-Term Visual Localization with Hybrid Features} 

\titlerunning{Visual localization with dense semantic map and hybrid features}
%
\author{Tianxin Shi\inst{1,2} \and
Hainan Cui\inst{1,2} \and
Zhuo Song\inst{1,2} \and
Shuhan Shen\inst{1,2} }
\authorrunning{Shi et al.}
%
\institute{NLPR, Institute of Automation, Chinese Academy of Sciences \and
University of Chinese Academy of Sciences
}
\maketitle

\begin{abstract}
Visual localization plays an important role in many applications. However, due to the large appearance variations such as season and illumination changes, as well as weather and day-night variations, it’s still a big challenge for robust long-term visual localization algorithms. In this paper, we present a novel visual localization method using hybrid handcrafted and learned features with dense semantic 3D map. Hybrid features help us to make full use of their strengths in different imaging conditions, and the dense semantic map provide us reliable and complete geometric and semantic information for constructing sufficient 2D-3D matching pairs with semantic consistency scores. In our pipeline, we retrieve and score each candidate database image through the semantic consistency between the dense model and the query image. Then the semantic consistency score is used as a soft constraint in the weighted RANSAC-based PnP pose solver. Experimental results on long-term visual localization benchmarks demonstrate the effectiveness of our method compared with state-of-the-arts.

\end{abstract}

\section{Introduction}

Visual localization, which has recently drawn a lot of attentions, is estimating the 6DOF camera pose of an image with respect to a 3D model of the scene. It plays a central role in many applications, such as augmented reality \cite{castle2008video}, Structure-from-Motion (SfM) \cite{schonberger2016structure}, loop closure detection \cite{7989618}, re-localization \cite{li2010location} in SLAM \cite{mur2015orb}, and autonomous vehicles.

There are generally three types of visual localization, named image-based methods \cite{arandjelovic2016netvlad,lowry2016visual,sattler2016large,torii201524}, learning-based methods \cite{cao2013graph,chen2017deep,kendall2015posenet}, and structure-based methods \cite{liu2017efficient,sattler2017efficient,svarm2017city,zeisl2015camera}. Compared to the former two types, structure-based approaches tend to provide more accurate camera poses. Although traditional structured-based localization methods work well when the query and database images are taken under similar conditions, they tend to fail when the scene has large appearance variations such as different season, illumination, or weather. The reason is that these methods rely heavily on local feature descriptors which are very sensitive to changes in appearance, and may produce too many matching outliers in long-term scenarios which will lead to the failure of visual localization.

To address the above problems, recently learning-based features \cite{detone2018superpoint,dusmanu2019d2,luo2019contextdesc,revaud2019r2d2} are introduced into the visual localization community to improve the localization performance under large appearance changes or in weakly textured scenes. There are mainly two types of learned features, one is the \textbf{detect-then-describe} approaches \cite{luo2019contextdesc,luo2018geodesc,tian2017l2,tian2019sosnet} which rely on handcrafted detectors, like SIFT \cite{lowe2004distinctive} and use learned descriptor to replace the handcrafted descriptor on the same patch, the other is the \textbf{detect-and-describe} approaches \cite{detone2018superpoint,dusmanu2019d2,revaud2019r2d2} which use end-to-end learning method to generate both feature locations and descriptions. Though learning-based features have achieved good results on benchmarks \cite{sattler2018benchmarking}, traditional handcrafted features, like SIFT, still play an important role in real-world applications due to its high detection accuracy and independence of training data. In this paper, we combine the handcrafted and learned features to use their advantages synthetically. In this way, the commonly-used sparse 3D model generated from SfM cannot be used anymore, especially for the detect-and-describe type features whose feature localization are totally different from SIFT. Although re-triangulation could be performed to generate another sparse 3D model for detect-and-describe descriptors, the accuracy of the re-triangulated model is less accurate than that of the SfM model computed from SIFT because the feature location accuracy of detect-and-describe approaches cannot compete with the SIFT detector \cite{lowe2004distinctive}. To solve this problem, in this paper we propose using dense 3D model to replace the sparse SfM model using off-the-shelf Multiple View Stereo (MVS) method in order to adapt to various types of features.

Besides the learned features, incorporating high-level image semantics into the geometrical localization process is another effective way to improve the visual localization robustness \cite{8578819,toft2017long,toft2018semantic}. Inspired by these methods, we further upgrade the dense 3D model to semantic dense model. Compared to the sparse semantic model used in \cite{toft2018semantic}, the dense semantic model has more 3D points that could be used for semantic consistency check, which leads the semantic consistency score for each retrieved image to be more discriminative and could help us to pick more likely correct retrieved database images.

Compared to recent visual localization algorithms, this paper makes following contributions: 

\begin{itemize}
    \item We combine handcrafted and learned features such that we can make full use of their strengths in different viewing conditions.
    \item We assign each retrieved database image a semantic consistency score as a soft constraint to help us to pick more likely correct candidate database images.
    \item We propose to use dense semantic 3D map to adapt to hybrid descriptors and improve the discrimination of semantic consistency scores.
\end{itemize}

The rest of the paper is organized as follow: Sec. \ref{2} introduces related work. Sec. \ref{3} gives the pipeline and details each step of our method. Sec. \ref{4} shows the experiments on long-term visual localization benchmarks. Sec. \ref{5} draws a conclusion.

\section{Related work}
\label{2}

In this section we reviewed several topics that are closely related to this paper, including different types of visual localization methods, local features, localization with semantics, and the localization benchmarks.

\noindent\textbf{2D image-based localization}. This type of approaches \cite{arandjelovic2016netvlad,lowry2016visual,sattler2016large,torii201524} solve visual localization as an image retrieval problem. It was mainly used for place recognition and loop-closure in the past. In these methods, a set of most similar database images are acquired for the query image using image retrieval techniques. They use the pose of top-ranked retrieved image as the query pose or use top-$k$ ranked retrieved images to compute the pose for the query image. Retrieval techniques often based on some standard algorithm such as Bag-of-Words (BoW) representations \cite{sivic2003video} which are acquired from a vocabulary tree with inverted files, or other representations such as Fischer Vectors \cite{jegou2011aggregating} and VLAD \cite{arandjelovic2013all} which records the distance from the feature point to the nearest visual word.

As it is challenging for visual localization under large viewpoint changes and illumination variations, Torii et al. \cite{torii201524} propose a method that adding some synthetic images from novel viewpoints to the database images so that they make image retrieval more accurate and easier under such tough conditions. With the development of deep learning, Convolutional Neural Networks (CNNs) have been used for image retrieval. NetVLAD \cite{arandjelovic2016netvlad} is one of the most representative works which implements an end-to-end place recognition network. Even under large viewing condition variations such as day-night changes, this method could still have a good performance.

\noindent\textbf{3D structure-based localization}. This type of methods \cite{li2012worldwide,li2010location,liu2017efficient,sattler2017efficient,svarm2017city,zeisl2015camera} represent the scene as a 3D model, and assign each 3D point one or several visible local descriptors. Given a query image, they establish 2D-3D correspondences through descriptor matching. Then, they apply a PnP solver \cite{kneip2011novel} inside a RANSAC loop \cite{mach1981random} to recover the query image pose from these matches.

For handling more complex scenes, prioritized matching strategies \cite{li2010location,sattler2017efficient} were proposed focusing on accelerating descriptor matching. It was terminated if a certain number of matches have been found. Some approaches limit the searching space with co-visibility information \cite{li2012worldwide,liu2017efficient} or retrieval-based method \cite{irschara2009structure,sattler2015hyperpoints}, so that they only need to search in the part of the whole 3D points and obtain more correct matches thereby. For handling large amount of outliers, geometric outlier filtering methods \cite{svarm2017city,zeisl2015camera} determine how consistent one match is by using geometric reasoning.

\noindent\textbf{Learning-based localization}. Many approaches \cite{cao2013graph,chen2017deep,clark2017vidloc,kendall2017geometric,kendall2015posenet,walch2017image} have been proposed by utilizing deep learning. Some of them solve localization problems as place classification and CNNs are used for learning classifiers \cite{cao2013graph,chen2017deep}. Besides, some approaches \cite{clark2017vidloc,kendall2017geometric,kendall2015posenet,walch2017image} based on CNNs learn to regress 2D-3D matches or camera poses directly. However, such learning-based methods have not yet achieve a good performance compared with the state-of-the-arts.

\noindent\textbf{Local features}. For obtaining 2D-3D matches, we need to extract features from images. Traditionally, we use handcrafted feature descriptors, e.g., SIFT (Scale-Invariant Feature Transform) \cite{lowe2004distinctive}, SURF (Speeded Up Robust Features) \cite{bay2006surf}, ORB (Oriented FAST and Rotated BRIEF) \cite{rublee2011orb}, etc. These traditional feature descriptors are invariant to scale and perspective changes and are also adaptable to slight light changes and noises.

However, practical visual localization often need to face the challenge of large illumination and appearance variations, where traditional hand-crafted feature matching would tend to fail. Recently, a lot of CNN-based descriptor learning methods have been proposed. The detect-then-describe approaches, e.g., L2-Net \cite{tian2017l2}, HardNet \cite{mishchuk2017working}, GeoDesc \cite{luo2018geodesc}, ContextDesc \cite{luo2019contextdesc}, SOS-Net \cite{tian2019sosnet}, etc., first use SIFT
detector to obtain keypoints location, then produce $N \times N$ patches around the keypoints location and convert them to $D$ dimensional descriptors. Compared to detect-then-describe approaches, detect-and-describe methods, e.g., SuperPoint \cite{detone2018superpoint}, D2-Net \cite{dusmanu2019d2}, R2D2 \cite{revaud2019r2d2}, etc., training detector and descriptor jointly, first compute a set of feature maps which are then used to detect keypoints and compute their corresponding descriptors. These CNN-based learning descriptors do have better performance than traditional hand-crafted descriptors on the challenging long-term visual localization benchmarks.

\noindent\textbf{Localization with semantics}. The idea of taking advantage of semantics is not new. Stenborg et al. \cite{stenborg2018long} propose a particle filter-based semantic localization method without the need for traditional local feature descriptors (SIFT, SURF, etc.). Toft et al. \cite{toft2018semantic} assign each 2D-3D match a semantic consistency score for weighted RANSAC-based PnP solver by projecting 3D semantic points into the query semantic segmented images. Sch{\"o}nberger et al. \cite{8578819} proposed training a network for obtaining descriptors and completing the semantic scene by utilizing geometric and semantic information jointly. Descriptors are used to establish 3D-3D matches and estimate an alignment as the query pose. Toft et al. \cite{toft2017long} refined pose estimation by optimizing the semantic consistency of the curve segments and the projected 3D points.

\noindent\textbf{Localization benchmarks}. Query images in some datasets were taken in the same condition as database image so that these datasets are unchallenging and unpractical. Even the well-known datasets like KITTI \cite{geiger2013vision} and North Campus Long-Term (NCLT) \cite{carlevaris2016university} do not contain strong illumination or large viewpoint changes.

Recently, Sattler et al. \cite{sattler2018benchmarking} created the benchmark for long-term visual localization. They manually labelled lots of 2D-3D matches for query images to overcome the impact of scene changes for obtaining the ground truth. In this benchmark, the RobotCar Seasons dataset contains lots of challenging conditions including illumination changes (dawn/noon), seasonal variations (winter/summer), as well as weather (dust/sun/rain/snow) and day-night changes. The main difficulties of Aachen Day-Night dataset are large viewpoint changes and localizing night-time images. The challenge of the Extended CMU Seasons dataset is the large variations in appearance of the scene, especially in the suburban and park regions, where appearance changes drastically under season and illumination variations.

\begin{figure}
\centering  
\includegraphics[width=12cm,height=5.5cm]{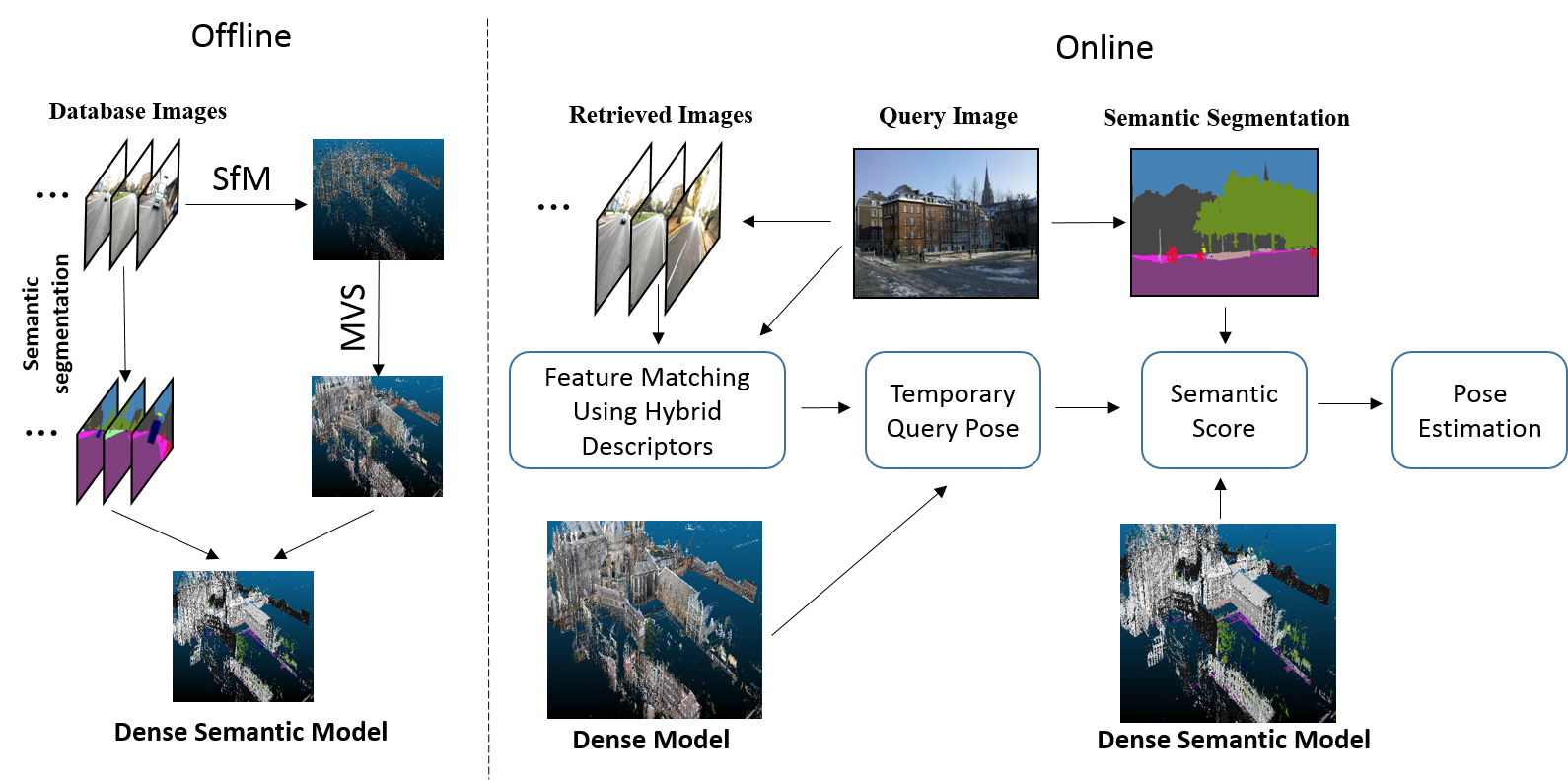}  
\caption{Pipeline of the proposed method} 
\label{fig:1}   
\end{figure}

\section{The Proposed Method}
\label{3}
Structured-based visual localization approaches rely heavily on local feature descriptors for finding enough correct 2D-3D matches between query image and 3D model, which are used to recover the query pose by applying a PnP solver. However, robust long-term visual localization is still a big challenge under the large appearance and viewing condition variants due to the lack of enough correct matches. In order to solve this problem, we proposed a novel visual localization method based on handcrafted and learned features with a dense semantic 3D map. The key idea of this paper comes from the following two observations.

The first observation is that in structure-based visual localization, handcrafted and learned features have their own advantages and applications. Handcrafted features, such as SIFT, SURF, ORB, etc., are invariant to scale and perspective changes and are also adaptable to slight light changes and noises, but they are too fragile to produce enough correct matches when facing large imaging condition changes. On the contrary, learned features have better performance to produce enough correct matches under difficult imaging conditions, like day-night changes, but their effectiveness somewhat depends on the training data. Besides, for end-to-end learned features, like SuperPoint, D2-Net, R2D2, etc., their 2D location accuracy is not as good as handcrafted competitors like SIFT, and may lead to a reduction in 3D positioning accuracy. Thus, it is viable to combine handcrafted and learned features together to make full use of their strengths in different imaging conditions.

The second observation is that compared to local features, high-level image semantic segmentation is an invariant scene representation which should not be affected by seasonal or other changes, and has begun to play an important role in visual localization. The semantic constraint is usually measured by first projecting visible 3D points on the query image, then checking the label consistency between 3D point in the map and its 2D projection in image. As a result, the more complete and dense the 3D model is, the more discriminative the measurement of semantic consistency is.

Based on these two observations, we proposed to use dense semantic 3D map to incorporate hybrid features and semantics into the structured-based visual localization pipeline. There are two benefits of using dense 3D map. One is that different types of features could use the same dense 3D map without creating a sparse 3D map for each kind of features. The other benefit is that the dense semantic map could produce more discriminative semantic consistency scores than the sparse model, and help us to pick more likely correct retrieved database images. In the following subsections, we describe our overall pipeline, and detail each step.

\subsection{Pipeline}
\label{3.1}

The pipeline of the proposed method is shown in Fig. \ref{fig:1}. First we use off-the-shelf SfM \cite{openMVG,schonberger2016structure} and MVS \cite{bleyer2011,schonberger2016,shen2013} algorithms to build a dense model of the scene using all database images, and extract handcrafted and learned features for each image. Then the dense 3D model is enhanced as a dense semantic map using image semantic segmentation results. For each query image, we get a set of candidate database images based on image retrieval, and establish 2D-3D matches through each pair of query and retrieved image using hybrid descriptors. Then we project all visible semantic 3D points into the query image through the recovered temporary pose, and measure the consistent semantic label number of 3D points and their projections on the query image. Finally, we use the semantic consistent number as a soft constraint in the weighted RANSAC-based PnP solver.

\begin{figure}
\centering  
\includegraphics[width=10cm,height=6cm]{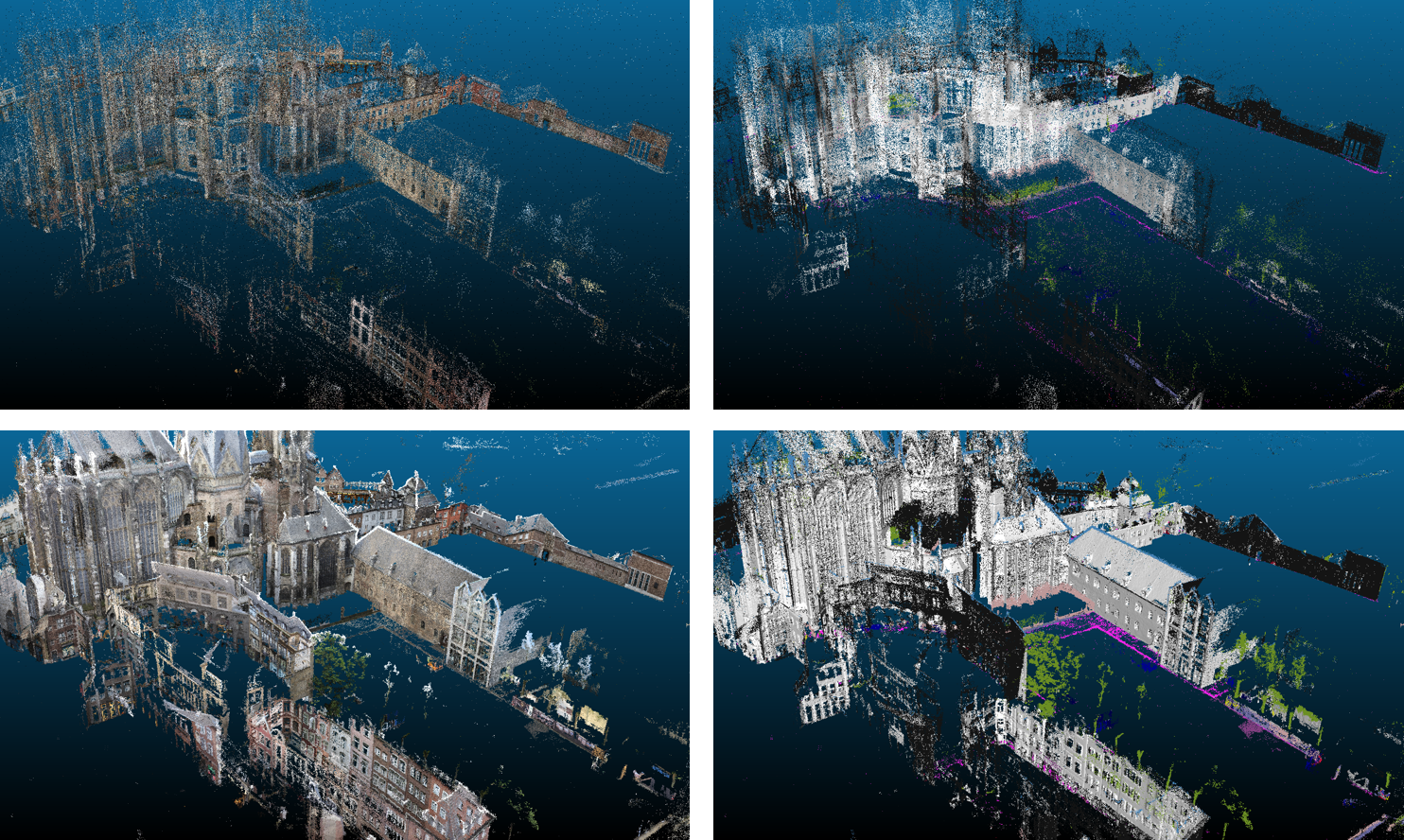}  
\caption{An example for different types of 3D model. The four images are taken from the same viewpoint. The upper left image shows the sparse model of the scene, while the lower left image represents the dense 3D model as comparison. And the upper and lower right images represent the sparse and dense model with semantics (different color represents different label).} 
\label{fig:2}   
\end{figure}

\subsection{Dense Semantic Map Construction}
\label{3.2}
For building a unified and accurate 3D model for all kinds of features, we propose to construct a dense semantic 3D map. The detailed building steps are described in the following. As for provided database images, we first perform semantic segmentation and run a SfM pipeline \cite{schonberger2016structure} to build a sparse 3D model of the scene. Given the calibrated images from SfM, we perform MVS to generate depth-map for each image and the merged dense point cloud. Among various geometry and learning based MVS methods, the PatchMatch-based MVS methods \cite{bleyer2011,schonberger2016,shen2013} are the top performing approaches on popular MVS benchmarks \cite{Knapitsch2017,Schps2017}. Here we use a mature PatchMatch-based MVS pipeline \cite{openMVS} to generate the dense map, which consists of neighbor views selection, propagation-based depth map computation, depth map filtering, and depth map merging. After that, each 3D point is assigned a semantic label by maximum voting of all reprojection pixel labels in all its visible images. According to the semantic label, we can remove unstable objects from the map, such as person, car, bus, sky, etc., which are noises for visual localization, and obtain a cleaner dense semantic 3D map (cf. Fig. \ref{fig:2}). Compared to sparse semantic model, the dense model has more 3D points could be used for semantic consistency check, which makes semantic consistency scores more differentiated. Note that PatchMatch-based MVS is a typical depth map merging based MVS method, i.e., the depth map for each image is computed during the MVS computation process. Thus for each feature point in the image, no matter handcrafted or learned one, its corresponding 3D point could be computed by its depth (if exist) in the depth map and the camera parameters.

\subsection{Feature Extraction}
\label{3.3}
In practice, visual localization methods should be robust to a wide range of viewing condition changes, such as illumination, weather and seasonal changes, as well as day-night changes and viewpoints changes. Until now, there are still lots of work tend to use typical handcrafted feature, e.g., SIFT, to extract local features for the tasks of 3D reconstruction and visual localization. SIFT descriptor is achieved by assembling a high-dimensional vector representing the image gradients within a local region of the image. It’s invariant to image scale and rotation, and is able to provide robust matching in most regular cases. However, when the viewing condition changes greatly, the SIFT detector becomes unstable due to the fact that low-level image information used by SIFT is often significantly more affected by the large appearance changes, which will cause the failure of feature matching so that we could not locate cameras successfully.

Recently, with fast developments of CNN-based features, the learning-based features began to show better performance in visual localization than traditional handcrafted features under challenging environments. These learned features use deep neural network to learn how to detect or how to describe the features in a data-driven way. Compared to SIFT which only considers small image regions, learned descriptors could use more information of image, e.g., colors, larger patches, higher-level structures, scene layout, etc., which makes them have better performance than SIFT on the task of feature matching under difficult imaging conditions. 

\begin{figure}
\centering  
\includegraphics[width=9cm,height=5cm]{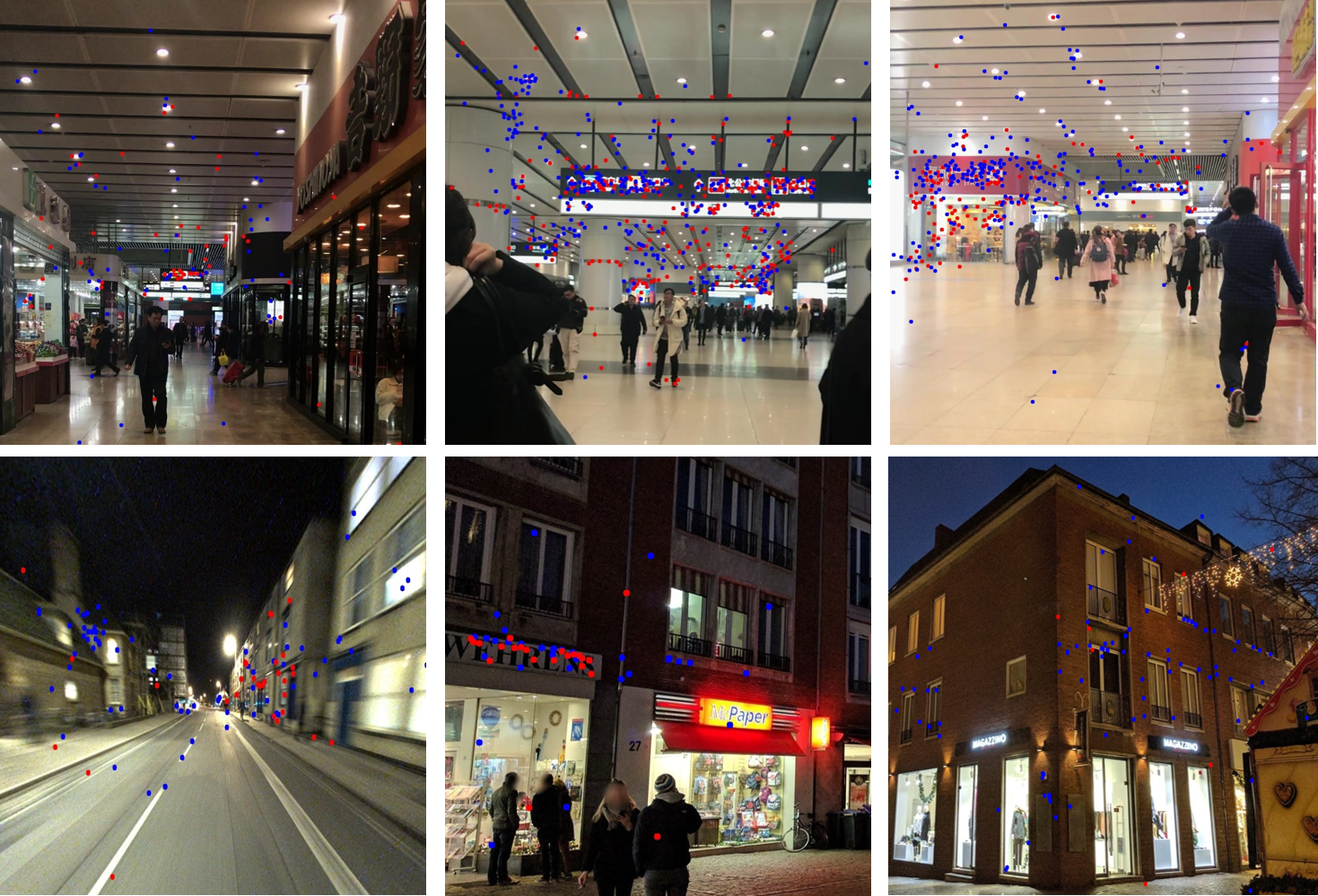}  
\caption{Examples of the inlier matches of SIFT (red) and R2D2 (blue) in six different scenes. They play a complementary role for each other, which makes the locations of all matching points more widely distributed on the image and finally could improve the visual localization accuracy.} 
\label{fig:3}   
\end{figure}

Based on the above observations, which learned feature should we choose in our pipeline becomes the first problem to be solved. We pick three notable learned descriptors, D2-Net, R2D2 and SuperPoint as candidates, which are open sourced and top-ranked on the benchmark of local feature challenge \cite{sattler2018benchmarking}. Based on our proposed pipeline, we test SIFT and these learned features respectively on the Aachen Day-Night dataset in the long-term visual localization benchmark \cite{sattler2018benchmarking}. The focus of this dataset is localizing day-time and night-time queries against a 3D dense model constructed from only day-time database images. From the result, we found based on the same dense semantic model, SIFT has higher localization accuracy in day condition, while R2D2 has better performance in night condition (details of the experiments are shown in Sec. \ref{4} Tab. \ref{tab1}). But in practice, it's hard to find a clear dividing line between day and night, so it's reasonable to use SIFT and R2D2 together to adapt to different lighting conditions. The experimental results also show that the combination does not eliminate each other's advantages, but can achieve state-of-the-art performance. Besides, we found with hybrid features the locations of matched points on the query image are more widely distributed than only using a single type of feature (cf. Fig. \ref{fig:3}), which shows these two types of features could play a complementary role to each other, so using them together is a reasonable and effective way in visual localization.

\subsection{Image Retrieval and Semantic Consistency Evaluation}
\label{3.4}

\begin{figure}
\centering  
\includegraphics[width=11cm,height=6cm]{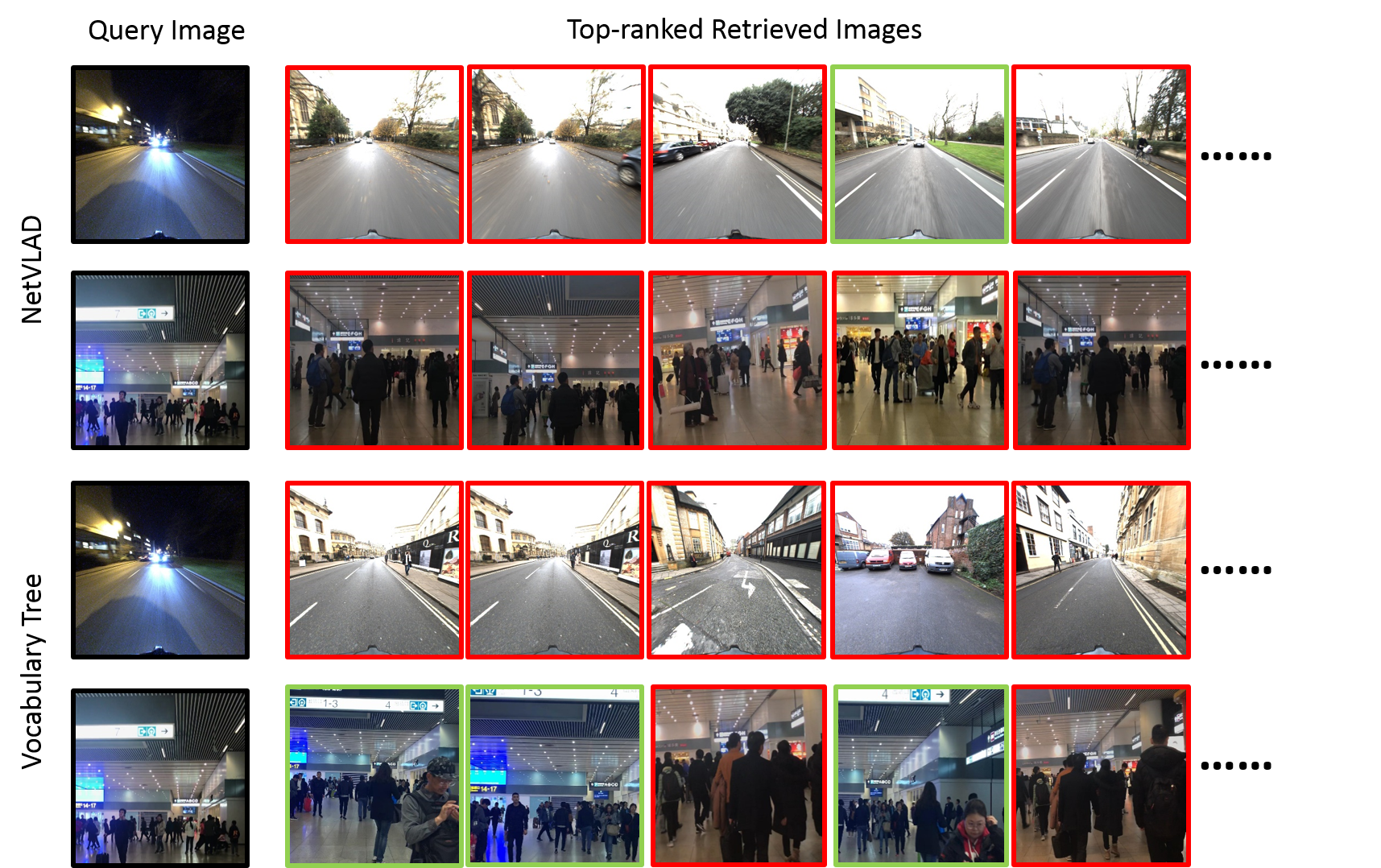}  
\caption{Illustration of the image retrieval results on two challenging scenes using NetVLAD and vocabulary tree respectively. Rows 1 and 3 show that NetVLAD outperforms vocabulary tree-based method under day-night changes, while rows 2 and 4 illustrate that vocabulary tree-based method is good at dealing with symmetrical and similar structures in the indoor scene. (images with green frames are inliers, and images with red frames are outliers)} 
\label{fig:4}   
\end{figure}

Given the 3D map and a query image, image retrieval techniques are first used to obtain a set of most similar database images. Here, we test two popular image retrieval approaches, NetVLAD \cite{arandjelovic2016netvlad} and the vocabulary tree-based method provided by Colmap \cite{schoenberger2016vote}. NetVLAD uses CNNs to learn the image representation end-to-end, while the vocabulary tree-based method use vocabulary tree to obtain the representations of image with spatial re-ranking. In the experiments, we found NetVLAD performs much better under large illumination variations, especially under day-night changes, but tends to have problems when facing the similar or symmetrical structures which often occurs in indoor scenes (cf. Fig. \ref{fig:4}). As we can see from rows 1 and 3 in Fig. \ref{fig:4}, under day-night changes, NetVLAD could obtain a correct image among the top-5 candidate images, while vocabulary tree-based method failed. From row 2 in Fig. \ref{fig:4} we can see that the first, second and fifth retrieved images are structurally symmetric compared to the query image, while the third and fourth retrieved images are structurally similar to the query image, but are not in the same place. The main reason is probably because NetVLAD learns the high-level image structure information and its base architecture was pretrained for ImageNet classification, which contains lots of data augmentation. As a result, the structurally similar or symmetrical images may obtain essentially the same feature representations from the output of CNN layers. 

Based on the above results, in order to increase the percentage of correctly recognized queries (Recall on all query images), an intuitive way is to use NetVLAD and vocabulary tree-based method together. However, we found using hybrid image retrieval methods would reduce the percentage of correctly retrieved images (Precision on one query image) for most conditions than using NetVLAD alone, which results in a decrease in localization accuracy. Therefore, we only choose NetVLAD as the image retrieval method in our pipeline, but we consider the issue of symmetry is a problem that needs special attention for learning-based retrieval method in the future.


After obtaining the top-ranked retrieved database images, we use SIFT and R2D2 together for feature matching between the query and retrieved images. Using one retrieved image at a time, we first compute 2D-2D matches and then recover their 3D coordinates according to the depth value provided by its depth map (the depth map of each database image is an intermediate product in MVS) to obtain the 2D-3D matches between the query image and 3D model. These 2D-3D matches are used to recover a temporary pose for the query image by applying a PnP solver. Then, we project all visible 3D points into the query image by its temporary pose to check the semantic consistency between the 2D-3D correspondences using the semantic 3D map and the query image's semantic segmentation result. Before projecting 3D points, we need to handle the situations of occlusion. Similar to \cite{toft2018semantic}, we only consider the 3D point that satisfies the following constraints:

\begin{equation}
\label{equ:1}　
d_{min}<\Vert \textbf{v}\Vert<d_{max} ,  \angle(\textbf{v},\textbf{v}_{m})<\theta   .
\end{equation}

\begin{equation}
\label{equ:2}　
\textbf{v}=\textbf{C}_{Q}-X , \textbf{v}_{m}=\frac{1}{2}(\textbf{v}_{l}+\textbf{v}_{u})   .
\end{equation}

\begin{figure}
\centering  
\includegraphics[width=6cm,height=4.5cm]{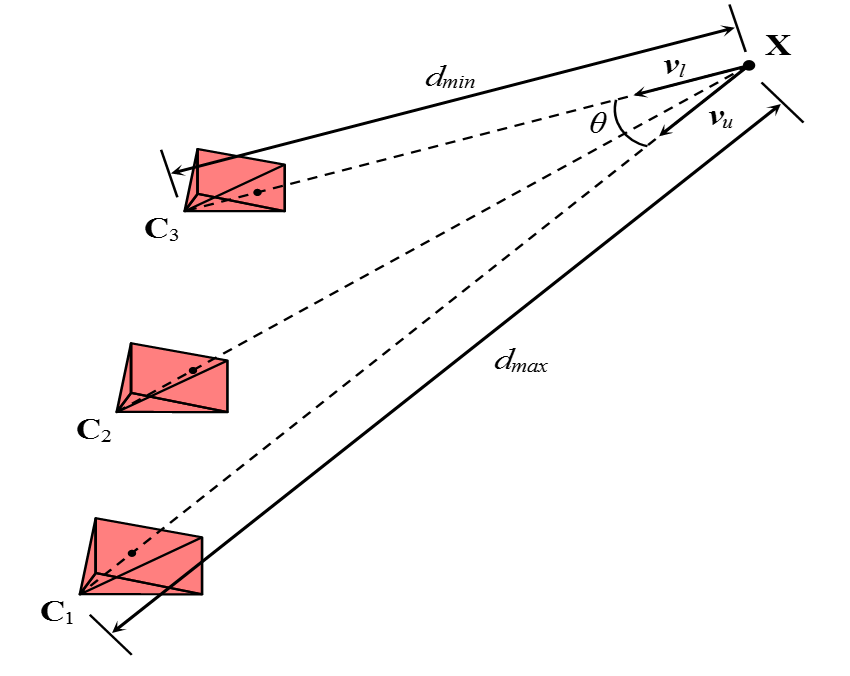}  
\caption{Illustration of the visibility angle and distance of a 3D point. $X$ is the 3D point, $\textbf{C}_1$, $\textbf{C}_2$ and $\textbf{C}_3$ are three visible database cameras, $d_{min}$ and $d_{max}$ are the minimum and maximum visible distances of $X$ respectively, and $\theta$ is the visible angle of $X$.} 
\label{fig:5}   
\end{figure}

where $\textbf{C}_{Q}$ is the camera center of the query image, $X$ is the 3D point, $d_{min}$ denotes the minimum distance between $X$ and its visible database cameras' centers, while $d_{max}$ denotes the maximum distance on the contrary. $\theta$ is the angle between these two extreme lines of sight $\textbf{v}_{l}$ and $\textbf{v}_{u}$, and an illustration of these variables are shown in Fig. \ref{fig:5}. This means the 3D point used for projection should be seen by the query image from similar distance and direction as this 3D point's visible database images.

We check the semantic consistent number of 3D points and their reprojections on the query image. The number is used as the semantic consistency score for this retrieved database image. For a negative retrieved image, the computed temporary pose of the query image is erroneous, so the semantic consistency score should be low, especially when we using dense semantic map with a lot of scene details. In this way, we could use the score to measure the correctness of the retrieved image, and use it as a soft constraint in the following pose estimation process.

\subsection{Pose Estimation}
\label{3.5}
After each retrieved database image is assigned a semantic consistency score, all 2D-3D matches produced by this retrieved image are assigned the same score. Finally, we use 2D-3D matches produced by all retrieved images with their semantic consistency score as weight in the RANSAC pose estimation process. More specifically, the semantic score of each match is normalized as a weight \textit{w} by the sum of all the scores of all matches, which means a 2D-3D match will be selected with probability \textit{w} inside the RANSAC loop. Finally, we run the weighted RANSAC-based PnP solver to recover the final query pose.

\section{Experimental Evaluation}
\label{4}
In this section we make detailed comparison of a set of start-of-the-art learned features and evaluate our proposed method on the long-term visual localization benchmarks \cite{sattler2018benchmarking}.

The benchmark \cite{sattler2018benchmarking} has provided sparse SfM model generated by Colmap \cite{schonberger2016structure} for each dataset. Using the SfM model and calibrated camera parameterts as inputs, we use openMVS \cite{openMVS}, which is a mature PatchMatch-based MVS system, to generate a dense 3D point cloud as well as depth map for each image. We should note that PatchMatch-based MVS has a depth map filtering step in which each depth $d$ in the raw depth map is projected on its neighbor views, and the depth consistency is measured to filter unreliable depth. More specifically, we back project $d$ to 3D then re-project it to its neighbor views, if the projected depth $d_R$ is consistent with the depth $d_N$ in the neighbor depth map, i.e. $(d_R-d_N)/d_N<\tau$ ($\tau$ is usually set to 0.01) on at least $N$ neighbor views, we say this point is stable, otherwise it is removed from the depth map. A big $N$ may result in a clean but incomplete depth map, on the contrary a small $N$ may cause the depth map to be noisy but more complete. $N$ is usually set to 2 or 3 in 3D reconstruction, however, in the long-term visual localization scenarios, the completeness of a depth map, which means more database image features have 3D information, usually have a big impact for the success of localization, and we found setting $N=1$ could achieve the best performance in all experiments.

For the semantic information, we use DeepLabv3+ network \cite{deeplabv3plus2018} with pre-trained \textit{xception65\underline{\hspace{0.5em}}cityscapes} model to segment all images and assign each dense 3D point a semantic label by maximum voting using this point's visible images. Additionally, for improving the semantic segmentation performance on night-time images, we manually annotate 30 night condition images from the origin RobotCar dataset \cite{maddern20171} and use them to fine-tune the pre-trained model. We use the same semantic classes as Cityscapes \cite{cordts2016cityscapes}.

In the image retrieval step, we use NetVLAD \cite{arandjelovic2016netvlad} with the pre-trained \textit{Pitts30K} model to generate 4096-dimensional descriptor vectors for each query and database image. Then, normalized L2 distances of the descriptors are computed, the top-$k$ best matched database images are chosen as candidate images. We set the retrieval number $k$ to 20 in day conditions and 30 in night conditions.

The evaluation criterion follows the measurement method from \cite{sattler2018benchmarking}. We measure the pose accuracy by recording percentage of query images which are localized within position error $X$m and orientation error $ Y^{\circ}$ compared to ground truth poses. The position error is evaluated by the Euclidean distance $\left\|C_{est}-C_{gt}\right\|_2$, where $C_{est}$ means the estimated query position and $C_{gt}$ means the ground truth position. The absolute orientation error $\left|\alpha\right|$ is computed from $2cos(\alpha)=trace(R_{gt}^{-1}R_{est})-1$, where $R_{gt}$ and $R_{est}$ represent the estimated and the ground truth camera rotation matrix respectively. In our experiments, two dataset from the benchmark, RobotCar Sensons and Aachen Day-Night, are used, and for measuring the pose accuracy under different thresholds, the intervals (0.25m, $ 2^{\circ}$), (0.5m, $ 5^{\circ}$), (5m, $ 10^{\circ}$) are used for RobotCar and Aachen day-time, and (0.5m, $ 2^{\circ}$), (1m, $ 5^{\circ}$), (5m, $ 10^{\circ}$) are used for Aachen night-time.

\begin{table}
\small
\caption{Comparison of different features on Aachen Day-Night dataset}
\centering
\resizebox{7cm}{1.3cm}{
\begin{tabular}{|c|c|c|}
\hline
 &day &night \\
 
\cline{2-3}
m & { .25 / .50 / 5.0} & { .50 / 1.0/ 5.0}\\
deg & { 2 / 5 / 10} & { 2 / 5/ 10}\\
\hline

{ D2-Net } & { 75.6 / 86.3 / 95.0} & { 40.8 / 66.3 / 85.7}  \\

\hline
{   SuperPoint   } & { 79.9 / 89.4 / 95.8} & { 41.8 / 64.3 / 83.7} \\

\hline
{ R2D2 } & { 83.0 / 91.5 / 96.5} & { 43.9 /\textbf{67.3} / 86.7} \\

\hline
{SIFT} & { \textbf{89.7} / 95.1 / \textbf{97.7}} & { 41.8 / 63.3 / 81.6} \\
\hline

{ SIFT + R2D2 } & { 89.3 / \textbf{95.4} / 97.6} & { \textbf{44.9} / \textbf{67.3} / \textbf{87.8}}  \\
\hline
\end{tabular}
\label{tab1}}
\end{table}

\subsection{Performance of Handcrafted and Learned Features}

We first evaluate three open sourced learned features, i.e., SuperPoint \cite{detone2018superpoint}, D2-Net \cite{dusmanu2019d2} and R2D2 \cite{revaud2019r2d2}, which are top ranked on the local feature challenge \cite{sattler2018benchmarking}. We use RootSIFT \cite{lowe2004distinctive} as baseline. For the fairness of the comparison, we use the same image retrieval results and the same dense model. In the experiment, for obtaining the best localization performance of each feature, we focus on some key parameters in feature matching, including whether to use mutual nearest neighbors validation and the using of different ratio thresholds of Lowe’s ratio test. As a result, we suggest using mutual nearest neighbors validation for all learned descriptors and RootSIFT. As for SuperPoint, we use the implementation and the pre-trained model provided by the authors with default feature extraction parameters and the thresholds of 0.9 for ratio test is suggested. As for D2-Net, we use the pre-trained \textit{d2\underline{\hspace{0.5em}}tf} model with multiscale detection to extract features and we recommend not to use ratio test. As for R2D2, we use the pre-trained \textit{r2d2\underline{\hspace{0.5em}}WASF\underline{\hspace{0.5em}}N8\underline{\hspace{0.5em}}big} model with multiscale detection to extract and retain top-20000 features. The same as D2-Net, We suggest not to use ratio test for R2D2 in feature matching step.

Table \ref{tab1} shows the comparison results of different features on the Aachen Day-Night dataset. As can be seen, among four evaluated features, SIFT outperforms the other three learned features in all three precisions in day-time condition, while R2D2 gets best results in all precisions in night-time condition. From the results, we consider SIFT is more suitable for images that are captured under similar imaging condition with the database images, while the learned features are more suitable under large appearance variations. Finally, we choose R2D2 and SIFT as hybrid features and the results are shown in the last row in Table \ref{tab1}. Although the performance of hybrid features is slightly inferior to SIFT in high and coarse precisions of day-time condition, they are better than each single type of feature in all other conditions.

\begin{table}
\small
\caption{Comparison results on Aachen Day-Night dataset}
\centering
\resizebox{9cm}{1.7cm}{
\begin{tabular}{|c|c|c|}
\hline
 &day &night \\
 
\cline{2-3}
m & { .25 / .50 / 5.0} & { .50 / 1.0 / 5.0}\\
deg & { 2 / 5 / 10} & { 2 / 5 / 10}\\
\hline

{Ours} & { \textbf{89.3} / \textbf{95.4} / 97.6} & {  44.9 / \textbf{67.3} / \textbf{87.8}}  \\

\hline
{Ours (Without Semantic)} & { 80.5 / 89.4 / 93.9} & { 42.9 / 66.3 / 86.7} \\

\hline

{spd2ms} & { 88.2 / 93.2 / 97.0} & { 42.9 / 66.3 / 84.7} \\

\hline
{ NetVLAD(top-20) + D2-Net(multi-scale) } & { 84.8 / 92.6 / 97.5} & { 43.9 / 66.3 / 85.7} \\
\hline
{ONavi-H} & { 77.7 / 89.0 / 95.8} & { \textbf{49.0} / 65.3 / 85.7} \\
\hline
{RT-PA-IR+CRBNet} & { 76.2 / 93.0 / \textbf{98.9}} & { 41.8 / 64.3 / 85.7} \\
\hline
{UR2KID} & { 79.9 / 88.6 / 93.6} & { 45.9 / 64.3 / 83.7} \\
\hline
{IR30+ML} & { 77.1 / 90.5 / 97.5} & { 43.9 / 61.2 / 82.7} \\
\hline
\end{tabular}
\label{tab2}}
\end{table}

\subsection{Performance on Benchmarks}
In this section, we compare our method against some top ranked approaches on the long-term visual localization benchmarks \cite{sattler2018benchmarking}. As we can see in Table 2, our proposed method outperforms all state-of-the-art approaches on the Aachen Day-Night dataset, except a lower accuracy than RT-PA-IR+CRBNet in coarse precision of day condition and ONavi-H in high precision of night condition. Meanwhile, from rows 1 and 2 in Table 2, the ablation study shows by exploiting semantic information, our proposed method has a significant improvement. This clearly validates our idea of using hybrid features with dense semantic model for the task of visual localization. 

To further validate our method, we also evaluate our method on the RobotCar Seasons dataset. In the experiments, we found part of query images in RobotCar Seasons dataset were taken in the opposite direction from the database images, which causes image retrieval-based localization methods cannot retrieve correct database images for this part of query images so that we cannot locate these images successfully. Even so, as we can see from Table 3, our method could still achieve the first place on the benchmark at the time of our paper submission, which again demonstrate that the proposed method is effective and robust for different conditions.

\begin{table}
\large
\caption{Comparison results on RobotCar Seasons dataset}
\centering
\resizebox{12cm}{1.8cm}{
\begin{tabular}{|c|c|c|}
\hline
 &day all &night all \\
 
\cline{2-3}
m & { .25 / .50 / 5.0} & { .25 / 0.50 / 5.0}\\
deg & { 2 / 5 / 10} & { 2 / 5 / 10}\\
\hline

{Ours} & { 54.6 / \textbf{81.9} / 96.9} & { 14.8 / 33.0 / 51.3}  \\

\hline
{RT-PA-IR+CRBNet} & { 55.3 / 81.8 / \textbf{98.5}} & { 11.5 / 26.5 / 39.2} \\

\hline
{SIFT+IR50+SIG+FM+R70-85+MUL+RE3+GPNP+MERGE} & { \textbf{57.2} / 81.5 / 97.4 } & { 9.3 / 30.1 / 53.3} \\

\hline
{DenseVLAD + D2-Net} & { 54.5 / 80.0 / 95.3} & { \textbf{20.4} / 40.1 / 55.0} \\
\hline
{Visual Localization Using Sparse Semantic 3D Map} & { 54.5 / 81.6 / 96.7} & { 12.3 / 28.5 / 46.5} \\
\hline
{ToDayGan + NetVLAD + D2-Net} & { 52.2 / 80.1 / 95.9} & { 20.3 / \textbf{46.8} / \textbf{73.7}} \\
\hline
{Hierarchical-Localization (multi-camera when available)} & { 53.8 / 80.4 / 96.0} & { 11.2 / 27.7 / 49.1 } \\
\hline
{Semantic Match Consistency} & { 50.3 / 79.3 / 95.2} & { 7.1 / 22.4 / 45.3} \\
\hline

\end{tabular}
\label{tab3}}
\end{table}

\section{Conclusion}
\label{5}
In this paper, we present a key step towards accurate visual localization algorithm. In order to use handcrafted and learned features together to make full use of their strengths in different imaging conditions, we propose to use dense 3D map instead of the sparse one in structure-based visual localization pipeline. Besides, by exploiting the semantic information, we further upgrade the dense 3D map to semantic dense map which is used to measure the semantic consistency score of each retrieved database image so that we can pick more likely correct retrieved images. Compared to sparse 3D model, dense semantic model has more 3D points that could be used for semantic consistency check, which leads the semantic consistency score to be more discriminative.

Experiments on Aachen Day-Night and Robotcar Seasons datasets from the visual localization benchmark show that our method outperforms current state-of-the-art methods in challenging conditions. In the future, we would like to further improve the integrity and accuracy of the dense 3D model and optimize the computational efficiency to cope with massive amounts of data. In addition, we intend to improve the localization performance when facing structurally symmetric or similar scenes.

%
%
\bibliographystyle{splncs04}
\bibliography{egbib}

\begin{thebibliography}{10}
\providecommand{\url}[1]{\texttt{#1}}
\providecommand{\urlprefix}{URL }
\providecommand{\doi}[1]{https://doi.org/#1}

\bibitem{arandjelovic2016netvlad}
Arandjelovic, R., Gronat, P., Torii, A., Pajdla, T., Sivic, J.:
  \protect{NetVLAD: CNN architecture for weakly supervised place recognition}.
  In: Proceedings of the IEEE Conference on Computer Vision and Pattern
  Recognition. pp. 5297--5307 (2016)

\bibitem{arandjelovic2013all}
Arandjelovic, R., Zisserman, A.: All about vlad. In: Proceedings of the IEEE
  conference on Computer Vision and Pattern Recognition. pp. 1578--1585 (2013)

\bibitem{bay2006surf}
Bay, H., Tuytelaars, T., Van~Gool, L.: Surf: Speeded up robust features. In:
  European conference on computer vision. pp. 404--417. Springer (2006)

\bibitem{bleyer2011}
Bleyer, M., Rhemann, C., Rother, C.: Patchmatch stereo-stereo matching with
  slanted support windows. In: BMVC (2011)

\bibitem{cao2013graph}
Cao, S., Snavely, N.: Graph-based discriminative learning for location
  recognition. In: IEEE Conference on Computer Vision and Pattern Recognition.
  pp. 700--707 (2013)

\bibitem{carlevaris2016university}
Carlevaris-Bianco, N., Ushani, A.K., Eustice, R.M.: University of michigan
  north campus long-term vision and lidar dataset. The International Journal of
  Robotics Research  \textbf{35}(9),  1023--1035 (2016)

\bibitem{castle2008video}
Castle, R., Klein, G., Murray, D.W.: Video-rate localization in multiple maps
  for wearable augmented reality. In: Wearable Computers, 2008. ISWC 2008. 12th
  IEEE International Symposium on. pp. 15--22. IEEE (2008)

\bibitem{deeplabv3plus2018}
Chen, L.C., Zhu, Y., Papandreou, G., Schroff, F., Adam, H.:
  \protect{Encoder-Decoder with Atrous Separable Convolution for Semantic Image
  Segmentation}. In: European Conference on Computer Vision (2018)

\bibitem{chen2017deep}
Chen, Z., Jacobson, A., S{\"u}nderhauf, N., Upcroft, B., Liu, L., Shen, C.,
  Reid, I., Milford, M.: Deep learning features at scale for visual place
  recognition. In: Robotics and Automation, 2017 IEEE International Conference
  on. pp. 3223--3230. IEEE (2017)

\bibitem{clark2017vidloc}
Clark, R., Wang, S., Markham, A., Trigoni, N., Wen, H.: Vidloc: A deep
  spatio-temporal model for 6-dof video-clip relocalization. In: Proceedings of
  the IEEE Conference on Computer Vision and Pattern Recognition. pp.
  6856--6864 (2017)

\bibitem{cordts2016cityscapes}
Cordts, M., Omran, M., Ramos, S., Rehfeld, T., Enzweiler, M., Benenson, R.,
  Franke, U., Roth, S., Schiele, B.: The cityscapes dataset for semantic urban
  scene understanding. In: Proceedings of the IEEE conference on computer
  vision and pattern recognition. pp. 3213--3223 (2016)

\bibitem{detone2018superpoint}
DeTone, D., Malisiewicz, T., Rabinovich, A.: Superpoint: Self-supervised
  interest point detection and description. In: Proceedings of the IEEE
  Conference on Computer Vision and Pattern Recognition Workshops. pp. 224--236
  (2018)

\bibitem{7989618}
Dubé, R., Dugas, D., Stumm, E., Nieto, J., Siegwart, R., Cadena, C.:
  \protect{SegMatch: Segment based place recognition in 3D point clouds}. In:
  2017 IEEE International Conference on Robotics and Automation. pp. 5266--5272
  (May 2017). \doi{10.1109/ICRA.2017.7989618}

\bibitem{dusmanu2019d2}
Dusmanu, M., Rocco, I., Pajdla, T., Pollefeys, M., Sivic, J., Torii, A.,
  Sattler, T.: D2-net: A trainable cnn for joint description and detection of
  local features. In: Proceedings of the IEEE Conference on Computer Vision and
  Pattern Recognition. pp. 8092--8101 (2019)

\bibitem{geiger2013vision}
Geiger, A., Lenz, P., Stiller, C., Urtasun, R.: Vision meets robotics: The
  kitti dataset. The International Journal of Robotics Research
  \textbf{32}(11),  1231--1237 (2013)

\bibitem{irschara2009structure}
Irschara, A., Zach, C., Frahm, J.M., Bischof, H.: From structure-from-motion
  point clouds to fast location recognition. In: 2009 IEEE Conference on
  Computer Vision and Pattern Recognition. pp. 2599--2606. IEEE (2009)

\bibitem{jegou2011aggregating}
Jegou, H., Perronnin, F., Douze, M., S{\'a}nchez, J., Perez, P., Schmid, C.:
  Aggregating local image descriptors into compact codes. IEEE transactions on
  pattern analysis and machine intelligence  \textbf{34}(9),  1704--1716 (2011)

\bibitem{kendall2017geometric}
Kendall, A., Cipolla, R.: Geometric loss functions for camera pose regression
  with deep learning. In: Proceedings of the IEEE Conference on Computer Vision
  and Pattern Recognition. pp. 5974--5983 (2017)

\bibitem{kendall2015posenet}
Kendall, A., Grimes, M., Cipolla, R.: Posenet: A convolutional network for
  real-time 6-dof camera relocalization. In: Proceedings of the IEEE
  international conference on computer vision. pp. 2938--2946 (2015)

\bibitem{Knapitsch2017}
Knapitsch, A., Park, J., Zhou, Q.Y., Koltun, V.: Tanks and temples:
  benchmarking large-scale scene reconstruction. ACM Transactions on Graphics

\bibitem{kneip2011novel}
Kneip, L., Scaramuzza, D., Siegwart, R.: A novel parametrization of the
  perspective-three-point problem for a direct computation of absolute camera
  position and orientation. In: CVPR 2011. pp. 2969--2976. IEEE (2011)

\bibitem{li2012worldwide}
Li, Y., Snavely, N., Huttenlocher, D., Fua, P.: Worldwide pose estimation using
  3d point clouds. In: European conference on computer vision. pp. 15--29.
  Springer (2012)

\bibitem{li2010location}
Li, Y., Snavely, N., Huttenlocher, D.P.: Location recognition using prioritized
  feature matching. In: European Conference on Computer Vision. pp. 791--804.
  Springer (2010)

\bibitem{liu2017efficient}
Liu, L., Li, H., Dai, Y.: \protect{Efficient Global 2D-3D Matching for Camera
  Localization in a Large-Scale 3D Map}. In: 2017 IEEE International Conference
  on Computer Vision. pp. 2391--2400 (Oct 2017)

\bibitem{lowe2004distinctive}
Lowe, D.G.: Distinctive image features from scale-invariant keypoints.
  International journal of computer vision  \textbf{60}(2),  91--110 (2004)

\bibitem{lowry2016visual}
Lowry, S., S{\"u}nderhauf, N., Newman, P., Leonard, J.J., Cox, D., Corke, P.,
  Milford, M.J.: Visual place recognition: A survey. IEEE Transactions on
  Robotics  \textbf{32}(1),  1--19 (2016)

\bibitem{luo2019contextdesc}
Luo, Z., Shen, T., Zhou, L., Zhang, J., Yao, Y., Li, S., Fang, T., Quan, L.:
  Contextdesc: Local descriptor augmentation with cross-modality context. In:
  Proceedings of the IEEE Conference on Computer Vision and Pattern
  Recognition. pp. 2527--2536 (2019)

\bibitem{luo2018geodesc}
Luo, Z., Shen, T., Zhou, L., Zhu, S., Zhang, R., Yao, Y., Fang, T., Quan, L.:
  Geodesc: Learning local descriptors by integrating geometry constraints. In:
  Proceedings of the European Conference on Computer Vision (ECCV). pp.
  168--183 (2018)

\bibitem{mach1981random}
Mach, C.: Random sample consensus: a paradigm for model fitting with
  application to image analysis and automated cartography. Readings in Computer
  Vision pp. 726--740 (1981)

\bibitem{maddern20171}
Maddern, W., Pascoe, G., Linegar, C., Newman, P.: 1 year, 1000 km: The oxford
  robotcar dataset. The International Journal of Robotics Research
  \textbf{36}(1),  3--15 (2017)

\bibitem{mishchuk2017working}
Mishchuk, A., Mishkin, D., Radenovic, F., Matas, J.: Working hard to know your
  neighbor's margins: Local descriptor learning loss. In: Advances in Neural
  Information Processing Systems. pp. 4826--4837 (2017)

\bibitem{openMVG}
Moulon, P., Monasse, P., Marlet, R., Others: Openmvg.
  \url{https://github.com/openMVG/openMVG}

\bibitem{mur2015orb}
Mur-Artal, R., Montiel, J.M.M., Tardos, J.D.: \protect{ORB-SLAM: A versatile
  and accurate monocular SLAM system}. IEEE Transactions on Robotics
  \textbf{31}(5),  1147--1163 (2015)

\bibitem{openMVS}
openMVS: \url{https://github.com/cdcseacave/openMVS}

\bibitem{revaud2019r2d2}
Revaud, J., De~Souza, C., Humenberger, M., Weinzaepfel, P.: R2d2: Reliable and
  repeatable detector and descriptor. In: Advances in Neural Information
  Processing Systems. pp. 12405--12415 (2019)

\bibitem{rublee2011orb}
Rublee, E., Rabaud, V., Konolige, K., Bradski, G.: Orb: An efficient
  alternative to sift or surf. In: 2011 International conference on computer
  vision. pp. 2564--2571. Ieee (2011)

\bibitem{sattler2017efficient}
Sattler, T., Leibe, B., Kobbelt, L.: Efficient \& effective prioritized
  matching for large-scale image-based localization. IEEE Transactions on
  Pattern Analysis \& Machine Intelligence  \textbf{39}(9),  1744--1756 (Sept
  2017)

\bibitem{sattler2015hyperpoints}
Sattler, T., Havlena, M., Radenovic, F., Schindler, K., Pollefeys, M.:
  Hyperpoints and fine vocabularies for large-scale location recognition. In:
  Proceedings of the IEEE International Conference on Computer Vision. pp.
  2102--2110 (2015)

\bibitem{sattler2016large}
Sattler, T., Havlena, M., Schindler, K., Pollefeys, M.: Large-scale location
  recognition and the geometric burstiness problem. In: Proceedings of the IEEE
  Conference on Computer Vision and Pattern Recognition. pp. 1582--1590 (2016)

\bibitem{sattler2018benchmarking}
Sattler, T., Maddern, W., Toft, C., Torii, A., Hammarstrand, L., Stenborg, E.,
  Safari, D., Okutomi, M., Pollefeys, M., Sivic, J., et~al.: Benchmarking 6dof
  outdoor visual localization in changing conditions. In: Proceedings of the
  IEEE Conference on Computer Vision and Pattern Recognition. pp. 8601--8610
  (2018)

\bibitem{8578819}
Schönberger, J.L., Pollefeys, M., Geiger, A., Sattler, T.: \protect{Semantic
  Visual Localization}. In: 2018 IEEE/CVF Conference on Computer Vision and
  Pattern Recognition. pp. 6896--6906 (June 2018).
  \doi{10.1109/CVPR.2018.00721}

\bibitem{schonberger2016structure}
Schonberger, J.L., Frahm, J.M.: Structure-from-motion revisited. In:
  Proceedings of the IEEE Conference on Computer Vision and Pattern
  Recognition. pp. 4104--4113 (2016)

\bibitem{schonberger2016}
Sch{\"o}nberger, J.L., Zheng, E., Frahm, J.M., Pollefeys, M.: Pixelwise view
  selection for unstructured multi-view stereo. In: ECCV. Springer (2016)

\bibitem{schoenberger2016vote}
Sch\"{o}nberger, J.L., Price, T., Sattler, T., Frahm, J.M., Pollefeys, M.: A
  vote-and-verify strategy for fast spatial verification in image retrieval.
  In: Asian Conference on Computer Vision (ACCV) (2016)

\bibitem{Schps2017}
Sch{\"o}ps, T., Sch{\"o}nberger, J.L., Galliani, S., Sattler, T., Schindler,
  K., Pollefeys, M., Geiger, A.: A multi-view stereo benchmark with
  high-resolution images and multi-camera videos. CVPR  (2017)

\bibitem{shen2013}
Shen, S.: Accurate multiple view 3d reconstruction using patch-based stereo for
  large-scale scenes. IEEE transactions on image processing  \textbf{22}(5),
  1901--1914 (2013)

\bibitem{sivic2003video}
Sivic, J., Zisserman, A.: Video google: A text retrieval approach to object
  matching in videos. In: null. p.~1470. IEEE (2003)

\bibitem{stenborg2018long}
Stenborg, E., Toft, C., Hammarstrand, L.: Long-term visual localization using
  semantically segmented images. In: 2018 IEEE International Conference on
  Robotics and Automation (ICRA). pp. 6484--6490. IEEE (2018)

\bibitem{svarm2017city}
Sv{\"a}rm, L., Enqvist, O., Kahl, F., Oskarsson, M.: City-scale localization
  for cameras with known vertical direction. IEEE Transactions on Pattern
  Analysis and Machine Intelligence  \textbf{39}(7),  1455--1461 (2017)

\bibitem{tian2017l2}
Tian, Y., Fan, B., Wu, F.: L2-net: Deep learning of discriminative patch
  descriptor in euclidean space. In: Proceedings of the IEEE Conference on
  Computer Vision and Pattern Recognition. pp. 661--669 (2017)

\bibitem{tian2019sosnet}
Tian, Y., Yu, X., Fan, B., Wu, F., Heijnen, H., Balntas, V.: Sosnet: Second
  order similarity regularization for local descriptor learning. In:
  Proceedings of the IEEE Conference on Computer Vision and Pattern
  Recognition. pp. 11016--11025 (2019)

\bibitem{toft2017long}
Toft, C., Olsson, C., Kahl, F.: \protect{Long-term 3D localization and pose
  from semantic labellings}. In: IEEE International Conference on Computer
  Vision Workshops. vol.~2, p.~3 (2017)

\bibitem{toft2018semantic}
Toft, C., Stenborg, E., Hammarstrand, L., Brynte, L., Pollefeys, M., Sattler,
  T., Kahl, F.: Semantic match consistency for long-term visual localization.
  In: European Conference on Computer Vision. pp. 391--408. Springer (2018)

\bibitem{torii201524}
Torii, A., Arandjelovic, R., Sivic, J., Okutomi, M., Pajdla, T.: 24/7 place
  recognition by view synthesis. In: Proceedings of the IEEE Conference on
  Computer Vision and Pattern Recognition. pp. 1808--1817 (2015)

\bibitem{walch2017image}
Walch, F., Hazirbas, C., Leal-Taixe, L., Sattler, T., Hilsenbeck, S., Cremers,
  D.: Image-based localization using lstms for structured feature correlation.
  In: Proceedings of the IEEE International Conference on Computer Vision. pp.
  627--637 (2017)

\bibitem{zeisl2015camera}
Zeisl, B., Sattler, T., Pollefeys, M.: Camera pose voting for large-scale
  image-based localization. In: Proceedings of the IEEE International
  Conference on Computer Vision. pp. 2704--2712 (2015)

\end{thebibliography}
\end{document}